\documentclass[final]{colt2026}

\title[Open Problem: AdamW Under Heavy Tails]{Open Problem: Is AdamW Effective Under Heavy-Tailed Noise?}
\usepackage{times}
\usepackage{bm}

\newcommand{\R}{\mathbb{R}}
\newcommand{\E}{\mathbb{E}}
\newcommand{\F}{\mathcal{F}}
\newcommand{\x}{\mathbf{x}}
\newcommand{\g}{\mathbf{g}}
\newcommand{\m}{\mathbf{m}}
\newcommand{\vct}{\mathbf{v}}
\newcommand{\uvec}{\mathbf{u}}
\newcommand{\evec}{\mathbf{e}}
\newcommand{\bL}{\bm{L}}
\newcommand{\bsigma}{\bm{\sigma}}
\newcommand{\eps}{\varepsilon}
\newcommand{\norm}[1]{\left\lVert #1 \right\rVert}
\newcommand{\abs}[1]{\left|#1\right|}
\newcommand{\ip}[2]{\left\langle #1,#2 \right\rangle}
\newcommand{\diag}{\operatorname{diag}}
\newcommand{\GC}{\eqref{eq:gc}}
\newcommand{\HTN}{\eqref{eq:ht}}
\newcommand{\ALN}{\eqref{eq:alignment}}
\newcommand{\BND}{\eqref{eq:update-bound}}

\coltauthor{
 \Name{Dingzhi Yu}\textsuperscript{1} \Email{yudz@lamda.nju.edu.cn}\\
 \Name{Hongyi Tao}\textsuperscript{1} \Email{221220032@smail.nju.edu.cn}\\
 \Name{Yuanyu Wan}\textsuperscript{2} \Email{wanyy@zju.edu.cn}\\
 \Name{Luo Luo}\textsuperscript{3} \Email{luoluo@fudan.edu.cn}\\
 \Name{Lijun Zhang}\textsuperscript{1} \Email{zhanglj@lamda.nju.edu.cn}\\
 \addr \textsuperscript{1}Nanjing University \quad
 \addr \textsuperscript{2}Zhejiang University \quad
 \addr \textsuperscript{3}Fudan University
}

\begin{document}

\maketitle

\begin{abstract}%
AdamW is the de facto optimizer for training large language models (LLMs), yet the theory behind it still lives mostly in finite-variance regimes. This is increasingly unsatisfying, as empirical evidence indicates that stochastic gradient noise in LLM pretraining is typically heavy-tailed. Recent work shows that sign-based optimizers such as Lion and Muon achieve sharp heavy-tailed rates, and that AdaGrad can also converge under heavy-tailed noise. However, no rigorous convergence theory for AdamW has yet been established in this regime. Can AdamW converge under the same heavy-tailed assumptions, or does its second-moment accumulator create a genuine obstruction? We formulate this as an open problem, prove a positive weighted-metric benchmark, and give a corridor lower-bound mechanism showing how denominator memory can hide large gradients.
\end{abstract}


\section{Introduction}

Heavy-tailed gradient noise is ubiquitous in modern machine learning~\citep{simsekli19tailindex,zhang2020whyheavytail,gurbuzbalaban2021heavy,battash2024revisiting}. This phenomenon naturally arises in language modeling tasks~\citep{piantadosi2014zipf}, as verified by a fruitful line of empirical research~\citep{kunstner2023noise,kunstner2024heavytail,ahn2024linearattention,kunstner2025scaling,yadav2025provable}.~\citet{yu2026sign} further validate on nanoGPT~\citep{nanogpt} pretraining that heavy-tailed noise persists across coordinates and matrix blocks, and that the noise magnitude tracks the gradient magnitude in a way not captured by classical bounded-variance models. Since AdamW~\citep{kingma15adam,loshchilov2019adamw} is the main workhorse for LLM training, understanding AdamW under this heavy-tailed regime is a central theoretical question.

This heavy-tailed viewpoint is sharpened by the empirical success of sign-based optimizers. Lion~\citep{chen2023symbolic} and Muon~\citep{jordan2024muon,liu2025muon} have shown consistent gains over AdamW in many practical scenarios~\citep{zhaoICLR2025deconstructing,shah2025practical,wen2025fantastic,semenov2025benchmarking}, including large-scale LLM systems such as Kimi-K2, \mbox{GLM-5}, and DeepSeek-V4~\citep{kimi2025k2,zeng2026glm,deepseekai2026deepseekv4}.~\citet{yu2026sign} establish a sharp theory for this phenomenon, showing that under finite $p$th moment heavy-tailed noise with $p\in(1,2]$, sign-based vector and matrix optimizers attain the sharp rate of $O(T^{-(p-1)/(3p-2)})$, which is tight in $T$~\citep{liu2025nonconvex}.  This makes the missing AdamW baseline more important, not less. If AdamW satisfies the same guarantee, then the empirical advantage of Lion and Muon must come from other aspects that cannot be captured by classical convergence theory. If AdamW fails, then the theory would help explain why it may be dominated by Muon in heavy-tailed LLM training regimes.

The open problem is therefore deliberately narrow.  The finite-variance AdamW baseline has essentially been supplied by~\citet{li2025adamw}, who prove an $O(\sqrt d/T^{1/4})$ $\ell_1$-convergence rate under coordinate-wise bounded variance.  We ask what remains true after replacing this variance assumption by a finite $p$th moment condition.  Existing Adam-type convergence analyses usually rely on finite/affine variance, bounded gradients, or bounded coordinate ratios~\citep{reddi2018convergence,zhang2022Adam,Li23adam,hong2024on,ahn2024adamema,peng2025simple,li2025adamw}.  Those assumptions do not address the case where $\E[\abs{\g_{t,i}}^p]<\infty$ for $p<2$ but $\E[\g_{t,i}^2]$ may be infinite, where $\g_{t,i}$ is the $i$-th coordinate of the stochastic gradient $\g_t$. Notably,~\citet{pmlr-v267-chezhegov25a} show that Adam could suffer from worse convergence when the noise is heavy-tailed.  A positive or negative AdamW theorem in this regime should therefore expose how the optimizer handles heavy-tailed noise, not only repeat the finite-variance descent algebra.

\section{Setup\label{sec:setup}}

Let $f:\R^d\to\R$ be differentiable and lower bounded, and write $\Delta=f(\x_1)-\inf_{\x\in\R^d} f(\x)$.  At step $t$, AdamW receives a mini-batch $\{\g_t^b\}_{b=1}^B$ and uses $\g_t=B^{-1}\sum_{b=1}^B\g_t^b$.  With $\m_0=\vct_0=\mathbf{0}$,
\begin{equation}\label{eq:adamw}
  \m_t=\beta_1\m_{t-1}+(1-\beta_1)\g_t,\ 
  \vct_t=\beta_2\vct_{t-1}+(1-\beta_2)\g_t\odot\g_t,\ 
  \x_{t+1}=(1-\eta_t\lambda)\x_t-\frac{\eta_t\m_t}{\sqrt{\vct_t}+\eps_t}.
\end{equation}
The schedules $\eta_t$ and $\eps_t$ may absorb the usual bias corrections.  The primary open case is $\lambda=0$.

Classical stochastic analyses assume smoothness of $f$ and finite oracle variance, for example $\E[\norm{\g_t-\nabla f(\x_t)}^2\mid\F_{t-1}]<\infty$.  In that regime, recent AdamW analyses can control the deterministic AdamW terms and the stochastic error by second moments~\citep{li2025adamw}.  We instead focus on the heavy-tailed regime motivated by language-model training, where both the local curvature and the noise scale may depend on the current gradient magnitude.  Following~\citet{yu2026sign}, let $D(\x)=\diag(\bL_0+\bL_1\odot\abs{\nabla f(\x)})$, where $\bL_0,\bL_1\in\R_+^d$ are coordinate-wise curvature parameters.  For all pairs $\x,\x'$ satisfying the local step-size condition of the generalized model, assume
\begin{equation*}\tag{GC}\label{eq:gc}
  \norm{\nabla f(\x')-\nabla f(\x)}_{D(\x)^{-1}}
  \le
  \norm{\x'-\x}_{D(\x)} .
\end{equation*}
This is the vector version of the gradient-curvature (GC) condition used by~\citet{yu2026sign} in their matrix analysis, and it is closely related to AdaGrad and generalized-smoothness analyses under nonuniform curvature~\citep{faw2023beyond,Li2023GS,liu2025adagrad,yu2025mirror,liu2026adagradheavy}.  For $p\in(1,2],\bsigma_0\succ\mathbf{0},\bsigma_1\succeq\mathbf{0}$, the stochastic gradients are captured by the coordinate-wise heavy-tailed (HT) noise model below:
\begin{equation*}\tag{HT}\label{eq:ht}
  \E[\g_t^b\mid\F_{t-1}]=\nabla f(\x_t),\ 
  \E\left[\left|\g_{t,i}^b-\nabla_i f(\x_t)\right|^p\mid\F_{t-1}\right]
  \le \bsigma_{0,i}^p+\bsigma_{1,i}^p\left|\nabla_i f(\x_t)\right|^p,\  \forall i\in[d].
\end{equation*}
Here subscripts such as $g_{t,i}^b$ and $\nabla_i f(\x_t)$ denote the $i$th coordinate.  Figures~1 and~2 of~\citet{yu2026sign} provide direct empirical support for this generalized heavy-tailed model on language-model pretraining runs.  Our partial results below focus on the \emph{clean specialization} $\bL_1=\bsigma_1=\mathbf{0}$, where we write $\bL=\bL_0$, $\bsigma=\bsigma_0>\mathbf{0}$, and $D=\diag(\bL)\succ0$.  In this specialization, we assume \GC{} globally, which is the regime used in the derivations below. This standard simplification is already sufficient to isolate the AdamW-specific barriers, as the function class for $\bL_1=\bsigma_1=\mathbf{0}$ is a subset of the more general class with $\bL_1,\bsigma_1\succ\mathbf{0}$.

\begin{remark}
  The conditions \GC{} and \HTN{} follow~\citet{yu2026sign} and align with the geometry of sign-based methods~\citep{jiang2025improved,tao2026when,yu2026stosignsgd}. Since Adam can be viewed as a variance-adapted SignSGD variant~\citep{balles2018dissecting,kunstner2023noise,peng2025simple}, the two methods share related geometry~\citep{balles2020geometry,liu2026optimizer}; this motivates studying AdamW on the same class. One can also leverage other assumptions based on a different geometry, such as standard $\norm{\nabla f(\x')-\nabla f(\x)}_2\le L_2\norm{\x'-\x}_2$ and $\E[\norm{\g_t^b-\nabla f(\x_t)}_2^p\mid\F_{t-1}]\le\sigma_2^p$, but that would typically incur explicit dimensional factors. More importantly, regardless of which curvature and noise models are considered, the same AdamW-specific technical barriers identified later in this paper would still apply, and the open problem would still be whether AdamW can overcome those barriers to match the sign-based rates.
\end{remark}

\section{Open Problem}

\paragraph{Open problem.}
\emph{Under \GC{} and \HTN{}, determine whether the convergence of AdamW can match the rate of sign-based methods under heavy-tailed noise in~\eqref{eq:target}.  More concretely, for the AdamW update defined in~\eqref{eq:adamw}, and already for the core case $\lambda=0$, either prove a nonasymptotic $\ell_1$-stationarity upper bound with the same heavy-tailed rate as sign-based methods, or construct a lower-bound instance showing that no such matching guarantee is possible for AdamW.}

The comparison target is the following.  Under the same heavy-tailed model,~\citet{yu2026sign} prove that sign-based optimizers attain the rate $O(T^{-(p-1)/(3p-2)})$, which is tight in $T$~\citep{liu2025nonconvex}.  In the \emph{clean specialization} $\bL_1=\bsigma_1=\mathbf{0}$ from Section~\ref{sec:setup}, this corresponds to
\begin{equation}
  \frac1T\sum_{t=1}^T \E[\norm{\nabla f(\x_t)}_1]
  \le O\!\left((\Delta\norm{\bL}_1)^{\frac{p-1}{3p-2}}
  \norm{\bsigma}_1^{\frac{p}{3p-2}}(BT)^{-\frac{p-1}{3p-2}}\right).
  \label{eq:target}
\end{equation}
This target is intentionally ambitious.  It asks whether AdamW can theoretically keep pace with sign-based methods once the finite-variance assumption is removed.  A slower AdamW upper bound under only \HTN{} would still be progress, but it would not answer this comparison.  Conversely, a lower bound should identify an AdamW-specific obstruction rather than merely restating that arbitrary heavy-tailed noise is difficult.  The key requirement is that the theorem use only \HTN, not a canonical finite variance assumption or a strong boundedness assumption.

The following propositions are our partial results, proved in Appendix~\ref{app:derivations}.  They separate a positive weighted benchmark from Adam-specific obstructions to converting weighted progress into plain $\ell_1$ stationarity.

\begin{proposition}[A positive weighted-metric benchmark]\label{prop:weighted}
Consider AdamW with $\lambda=0$, constant $\eta_t=\eta$, $\eps_t\ge0$, $\beta_2=\beta_1\in(0,1)$, and the convention $0/0=0$.  Define
\[
  a_{t,i}=\frac{\abs{m_{t,i}}}{\sqrt{v_{t,i}}+\eps_t},
  \qquad
  S_t=\ip{a_t}{\abs{\nabla f(\x_t)}} .
\]
Under the \emph{clean specialization} of \GC{} and \HTN, with independent mini-batch samples, there are choices of $\eta$ and $\beta_1$ such that, in the large-horizon regime $T(1-\beta_1)\gtrsim1$,
\[
  \frac1T\sum_{t=1}^T\E[S_t]
  \le O\!\left((\Delta\norm{\bL}_1)^{\frac{p-1}{3p-2}}
  \norm{\bsigma}_1^{\frac{p}{3p-2}}(BT)^{-\frac{p-1}{3p-2}}\right).
\]
The hidden constant depends only on $p$.
\end{proposition}

Proposition~\ref{prop:weighted} is useful for understanding how sign alignment and finite $p$th moments can yield the target heavy-tailed rate.  For SignSGD and Lion, $a_t\equiv\mathbf{1}$, so $S_t=\norm{\nabla f(\x_t)}_1$, and the same proof template becomes a plain $\ell_1$ guarantee.  \textbf{However, it still does not solve the open problem.}  For Adam, the data-dependent weights $a_{t,i}$ may collapse after a heavy-tailed outlier.  The result also uses the equal-memory regime $\beta_2=\beta_1$, while practical AdamW usually has $\beta_2\gg\beta_1$.  The unresolved case is therefore exactly the unequal-memory regime where the denominator may remember outliers much longer than the numerator (cf. Proposition~\ref{prop:memory}).  The weighted guarantee isolates the missing step, which is to prove that Adam's self-normalized weights cannot hide a large gradient, or construct an instance where they do.

\section{Technical Barriers}

The next two propositions are barrier mechanisms isolated in this paper and proved in Appendix~\ref{app:derivations}. They should be read as mechanism statements rather than previously known lower bounds.  Proposition~\ref{prop:skew} shows why unbiasedness alone cannot give alignment, while Proposition~\ref{prop:memory} gives a corridor lower-bound mechanism for long-memory Adam schedules.

Let $\uvec_t=\m_t/(\sqrt{\vct_t}+\eps_t)$.  By the descent lemma implied by \GC, any proof with $\lambda=0$ must control
\[
  f(\x_{t+1})
  \le
  f(\x_t)-\eta_t\ip{\nabla f(\x_t)}{\uvec_t}
  +\frac{\eta_t^2}{2}\norm{\uvec_t}_D^2 .
\]
The smoothness term depends on the size of the normalized update and is not where heavy tails enter.  The heavy-tailed difficulty is the alignment term.  Summing the descent inequality can only give a plain $\ell_1$ stationarity bound if the cumulative inner product $\sum_{t=1}^T\eta_t\ip{\nabla f(\x_t)}{\uvec_t}$ dominates $\sum_{t=1}^T\eta_t\norm{\nabla f(\x_t)}_1$.  We therefore isolate the \emph{exact alignment deficit quantity} as
\[
  \mathcal E_T(c)
  =
  \sum_{t=1}^T\eta_t\,
  \E\!\left[
  \left(c\norm{\nabla f(\x_t)}_1
  -\ip{\nabla f(\x_t)}{\uvec_t}\right)_+
  \right].
\]
By definition of $\mathcal E_T(c)$, it holds that
\begin{equation*}\tag{A}\label{eq:alignment}
  \begin{aligned}
  \sum_{t=1}^T\eta_t\E[\ip{\nabla f(\x_t)}{\uvec_t}]
  &\ge
  c\sum_{t=1}^T\eta_t\E[\norm{\nabla f(\x_t)}_1]
  -\mathcal E_T(c).
  \end{aligned}
\end{equation*}
Thus \ALN{} is a bookkeeping identity, not an extra assumption.  If $\mathcal E_T(c)$ is lower order, then the descent inequality gives \eqref{eq:target} up to the smoothness term.  The challenge is that, for AdamW, this deficit depends on the self-normalizer $\vct_t$ built from $\g_t^2$.  When $p<2$, $\E[\g_{t,i}^2|\F_{t-1}]$ may be infinite, so the finite-variance proof strategy of replacing $\vct_t$ by a conditional second-moment proxy is unavailable.  AdaGrad is informative only as a contrast: its monotone accumulator can be controlled under finite $p$th moments in recent work~\citep{liu2026adagradheavy}, whereas AdamW must additionally control the mismatch between an exponential numerator and an exponential squared-gradient denominator.
\begin{proposition}[Averaging is necessary]\label{prop:skew}
Fix any $\mu>0$, $p\in(1,2]$, and $q<1/2$.  At any one-dimensional point with true gradient $\mu$, there is an unbiased oracle distribution with finite $p$th noise moment such that AdamW with $\beta_1=\beta_2=0$ and $\eps_t=0$ has negative expected alignment, \(\E[\mu g/\abs{g}]=\mu(2q-1)<0\). Consequently, no AdamW theorem can prove \ALN{} from unbiasedness alone unless it explains how mini-batching, momentum, or robustness restores alignment.
\end{proposition}

\begin{remark}
  When $\beta_1=\beta_2=0$ and $\eps_t=0$, AdamW with $\lambda=0$ is exactly SignSGD.  Proposition~\ref{prop:skew} is therefore a warning about arbitrary unbiased heavy-tailed oracles, not about practical AdamW.
\end{remark}

\begin{proposition}[A corridor lower-bound mechanism]\label{prop:memory}
Fix $p\in(1,2)$ and consider 1D AdamW started from $x_1=0$, with $\lambda=0$, $\eps_t=0$, $\beta_1^2<\beta_2$, and $\rho_i=1-\beta_i$ with $\rho_2\le1/2$.  Let $C_\beta$ be the update bound in \BND{}.  There are universal constants $c,c_0,C_K,C_p>0$ such that the following holds.  For any $\mu,R,\sigma>0$ and $\delta\in(0,1/2)$, set
\[
  q=c_0\rho_2\log(1/\delta)\le1/2,\qquad
  M=\sigma q^{-1/p}\ge2\mu,\qquad
  K=\lceil C_K/\rho_2\rceil .
\]
There exists a $C^2$, lower-bounded one-dimensional function $f_{\mu,R}$ and an unbiased oracle satisfying
\[
  \E[\abs{g_t-f_{\mu,R}'(x_t)}^p\mid\F_{t-1}]\le 2\sigma^p
\]
such that, if $T>K$ and
\[
  \sum_{t=1}^{K}\frac{\eta_t C_\beta}{R}
  +
  \sum_{t=K+1}^{T}\frac{\eta_t}{R}
  \left[
  C_p\left(\frac{\mu}{\sigma}+\rho_1^{(p-1)/p}\right)
  \rho_2^{(2-p)/(2p)}
  (\log(1/\delta))^{1/p}
  +C_\beta\delta
  \right]
  \le c,
\]
then
\[
  \frac1T\sum_{t=1}^T\E[\abs{f_{\mu,R}'(x_t)}]\ge \mu/2 .
\]
Moreover $f_{\mu,R}$ has smoothness $L\asymp\mu/R$, initial gap $\Delta\asymp\mu R$, and hence $\Delta L\asymp\mu^2$.
\end{proposition}

Proposition~\ref{prop:memory} shows that the weighted guarantee in Proposition~\ref{prop:weighted} cannot be naively upgraded by proving that Adam's magnitude weights stay bounded away from zero.  Rare negative outliers with probability $q<1/2$ still leave SignSGD with positive drift, but Adam's $v_t$ remembers their squares and can make $\abs{m_t}/\sqrt{v_t}$ polynomially small.  For example, ignoring logarithms, take equal memory $\rho_1=\rho_2=T^{-a}$, $\eta_t/R\asymp \sqrt{\rho_2/T}$, and $\mu=\sigma T^{-s}$.  At the natural weighted-proof scale $a=p/(3p-2)$, one has $1/2<a<1$ for $p<2$, and there is a nonempty interval of $s<(p-1)/p$ for which Proposition~\ref{prop:memory} gives an $\Omega(\mu)$ stationarity lower bound, while the sign-method target \eqref{eq:target} is $o(\mu)$.  The result is not a universal lower bound over all AdamW hyperparameters, but it identifies a concrete long-memory regime where weighted stationarity can coexist with failure of plain $\ell_1$ stationarity.

\bibliography{ref}

\appendix

\section{Derivations Behind the Barriers}\label{app:derivations}

All derivations in this appendix use the \emph{clean specialization} $\bL_1=\bsigma_1=\mathbf{0}$, so that
\begin{align*}
  D(\x)=D=\diag(\bL_0)=\diag(\bL),\quad \bsigma=\bsigma_0>\mathbf{0}.
\end{align*}

\paragraph{Gradient-curvature implies descent.}
Let $s=\x'-\x$ and $D=\diag(\bL)=\diag(\bL_0)$.  Applying \GC{} to $\x+\tau s$ gives
\[
  \norm{\nabla f(\x+\tau s)-\nabla f(\x)}_{D^{-1}}
  \le \tau\norm{s}_{D}.
\]
Therefore, it holds that
\begin{equation*}
\begin{aligned}
  f(\x')-f(\x)-\ip{\nabla f(\x)}{s}
  =
  \int_0^1\ip{\nabla f(\x+\tau s)-\nabla f(\x)}{s}\,\mathrm{d}\tau
  \le
  \frac12\norm{s}_{D}^2 .
\end{aligned}
\end{equation*}
With $s=-\eta_t\uvec_t$, this gives the one-step inequality used before \ALN.  The decoupled weight-decay case follows with $s=-\eta_t(\uvec_t+\lambda\x_t)$ and extra terms involving $\lambda\x_t$.

\paragraph{Proof of \BND.}
When $\beta_1^2<\beta_2$, for one coordinate,
\[
  m_t=(1-\beta_1)\sum_{k=1}^t\beta_1^{t-k}g_k,\qquad
  v_t=(1-\beta_2)\sum_{k=1}^t\beta_2^{t-k}g_k^2 .
\]
Cauchy's inequality with weights $\beta_2^{t-k}$ gives
\[
  m_t^2\le
  (1-\beta_1)^2
  \left(\sum_{k=1}^t(\beta_1^2/\beta_2)^{t-k}\right)
  \left(\sum_{k=1}^t\beta_2^{t-k}g_k^2\right)
  \le
  \frac{(1-\beta_1)^2}{(1-\beta_1^2/\beta_2)(1-\beta_2)}v_t .
\]
Since $\eps_t\ge0$, this gives the deterministic update bound
\begin{equation*}\tag{B}\label{eq:update-bound}
  \abs{u_{t,i}}
  \le
  C_{\beta}=
  \frac{1-\beta_1}
  {\sqrt{(1-\beta_2)(1-\beta_1^2/\beta_2)}} .
\end{equation*}

\paragraph{Proof of Proposition~\ref{prop:weighted}.}
Let $\rho=1-\beta_1$, $\xi_t=\g_t-\nabla f(\x_t)$, and $\evec_t=\m_t-\nabla f(\x_t)$.  We use the von Bahr--Esseen inequality~\citep{vonBahr1965inequalities}.  For conditionally mean-zero random variables and $p\in[1,2]$, the $p$th moment of their sum is bounded, up to a universal constant, by the sum of their $p$th moments.  Applying it within a mini-batch gives
\[
  \E\!\left[\abs{\xi_{t,i}}^p\mid\F_{t-1}\right]
  \le O(\bsigma_i^p B^{1-p}).
\]
The momentum error decomposes as
\[
  \evec_t
  =
  -\beta_1^t\nabla f(\x_1)
  +\sum_{s=2}^t\beta_1^{t-s+1}
  \bigl(\nabla f(\x_{s-1})-\nabla f(\x_s)\bigr)
  +\rho\sum_{s=1}^t\beta_1^{t-s}\xi_s .
\]
By \GC{} and \BND{},
\[
  \norm{\nabla f(\x_s)-\nabla f(\x_{s-1})}_1
  \le
  \sqrt{\norm{\bL}_1}\norm{\nabla f(\x_s)-\nabla f(\x_{s-1})}_{D^{-1}}
  \le
  \eta C_\beta\norm{\bL}_1 .
\]
For the stochastic term, the weighted sequence
$\{\rho\beta_1^{t-s}\xi_{s,i}\}_{s=1}^t$ is a martingale difference sequence.  A second application of von Bahr--Esseen gives
\[
  \E\!\left[\norm{\rho\sum_{s=1}^t\beta_1^{t-s}\xi_s}_1\right]
  \le
  O\!\left(\frac{\norm{\bsigma}_1\rho^{1-1/p}}{B^{1-1/p}}\right),
\]
because $\rho^p\sum_{s=1}^t\beta_1^{p(t-s)}\le O(\rho^{p-1})$.
The initial term is controlled by smoothness and lower boundedness.
\[
  \norm{\nabla f(\x_1)}_1
  \le
  \sqrt{\norm{\bL}_1}\norm{\nabla f(\x_1)}_{D^{-1}}
  \le
  \sqrt{2\Delta\norm{\bL}_1}.
\]
Thus
\[
  \frac1T\sum_{t=1}^T\E[\norm{\evec_t}_1]
  \le
  \frac{\sqrt{2\Delta\norm{\bL}_1}}{\rho T}
  +\frac{\eta C_\beta\norm{\bL}_1}{\rho}
  +O\!\left(\frac{\norm{\bsigma}_1\rho^{1-1/p}}{B^{1-1/p}}\right).
\]
Now write $\uvec_t=a_t\odot\operatorname{sign}(\m_t)$.  Since $0\le a_{t,i}\le C_\beta$ by \BND{}, sign mismatch implies
\[
  \ip{\nabla f(\x_t)}{\uvec_t}
  \ge
  S_t-2C_\beta\norm{\evec_t}_1 .
\]
Combining this inequality with the descent lemma and $\norm{\uvec_t}_D^2\le C_\beta^2\norm{\bL}_1$ gives, after summing from $1$ to $T$,
\[
  \frac1T\sum_{t=1}^T\E[S_t]
  \le
  O\!\left(
  \frac{\Delta}{\eta T}
  +\frac{C_\beta^2\eta\norm{\bL}_1}{1-\beta_1}
  +\frac{C_\beta\norm{\bsigma}_1(1-\beta_1)^{1-1/p}}{B^{1-1/p}}
  +\frac{C_\beta\sqrt{\Delta\norm{\bL}_1}}{(1-\beta_1)T}
  \right).
\]
When $\beta_2=\beta_1$, we have $C_\beta=1$.  Choosing
\[
  \eta\asymp\sqrt{\frac{\Delta(1-\beta_1)}{\norm{\bL}_1T}},
  \qquad
  1-\beta_1
  \asymp
  \left(
  \frac{\sqrt{\Delta\norm{\bL}_1}B^{1-1/p}}
       {\norm{\bsigma}_1\sqrt{T}}
  \right)^{\frac{2p}{3p-2}}
\]
balances the first three terms.  Since $\bsigma>\mathbf{0}$, the initial-gradient term is lower order in the usual large-horizon, noise-dominated regime and can be absorbed into the same expression by enlarging the constant.  This gives Proposition~\ref{prop:weighted}.

\paragraph{Proof of Proposition~\ref{prop:skew}.}
Let $g=-a\mu$ with probability $1-q$ and $g=b\mu$ with probability $q$, where $b=(1+(1-q)a)/q$.  Then
\[
  \E[g]=\mu(-a(1-q)+bq)=\mu .
\]
The noise moment is finite:
\[
  \E[\abs{g-\mu}^p]
  =
  \mu^p\left((1-q)(a+1)^p+q(b-1)^p\right).
\]
For $\beta_1=\beta_2=0$ and $\eps_t=0$, AdamW uses $g/\abs{g}=\operatorname{sign}(g)$.  Hence
\[
  \E[\mu\,\operatorname{sign}(g)]
  =
  \mu\left((+1)q+(-1)(1-q)\right)
  =
  \mu(2q-1)<0 .
\]

\paragraph{Proof of Proposition~\ref{prop:memory}.}
First construct the corridor.  Let $h:\R\to[0,1]$ be a smooth cutoff with
$h(z)=0$ for $z\le-1$, $h(z)=1$ for $z\ge-1/2$, and $\abs{h'(z)}\le C_h$.
Set
\[
  f_{\mu,R}(x)=\mu R\int_{-\infty}^{x/R}h(z)\,\mathrm dz .
\]
Then $0\le f_{\mu,R}'(x)\le\mu$, $f_{\mu,R}'(x)=\mu$ for $x\ge-R/2$,
$f_{\mu,R}'(x)=0$ for $x\le-R$, and $\abs{f_{\mu,R}''(x)}\le C\mu/R$.
Thus $L\asymp\mu/R$, $\Delta=f_{\mu,R}(0)-\inf_x f_{\mu,R}(x)\asymp\mu R$, and
$\Delta L\asymp\mu^2$.

We next record the stopped-displacement implication.  For any trajectory
$x_{t+1}=x_t-\eta_tu_t$ started at $x_1=0$, let $u_t^+=\max\{u_t,0\}$ and
$D_T=\sum_{t=1}^T\eta_t\E[u_t^+]$.  Since only positive $u_t$ moves left along
the corridor,
\[
  (-x_t)_+\le \sum_{s<t}\eta_su_s^+ .
\]
Hence, if $D_T\le R/4$, Markov's inequality gives
\[
  \Pr(x_t<-R/2)\le \frac{2\E[(-x_t)_+]}{R}\le\frac12 .
\]
On the complementary event, $\abs{f_{\mu,R}'(x_t)}=\mu$, so
$T^{-1}\sum_{t=1}^T\E[\abs{f_{\mu,R}'(x_t)}]\ge\mu/2$.

It remains to upper bound Adam's expected positive displacement.  At a point with
$h_t=f_{\mu,R}'(x_t)\in[0,\mu]$, define the oracle by $g_t=h_t+\xi_t$, where
\[
  \xi_t=
  \begin{cases}
  -M, & \text{with probability }q,\\
  qM/(1-q), & \text{with probability }1-q.
  \end{cases}
\]
Then $\E[\xi_t\mid\F_{t-1}]=0$.  With $M=\sigma q^{-1/p}$ and $q\le1/2$,
\[
  \E[\abs{\xi_t}^p\mid\F_{t-1}]
  =
  qM^p+(1-q)\left(\frac{qM}{1-q}\right)^p
  \le 2\sigma^p .
\]

Let $K=\lceil C_K/\rho_2\rceil$, where $C_K$ is a sufficiently large absolute
constant, and let $\mathcal A_t$ be the event that at least one negative outlier
occurs in the last $K$ oracle calls.  Since
$q=c_0\rho_2\log(1/\delta)$ and $\rho_2\le1/2$, choosing $c_0C_K$ large enough
ensures $\Pr(\mathcal A_t^c)\le\delta$ for all $t>K$.  On $\mathcal A_t$, the
condition $M\ge2\mu$ implies that one of these gradients has magnitude at least
$M/2$.  Moreover $\beta_2^K$ is bounded below by an absolute constant, so
\[
  v_t\ge c\rho_2M^2 .
\]
For the numerator,
\[
  m_t
  =
  \rho_1\sum_{s=1}^t\beta_1^{t-s}f_{\mu,R}'(x_s)
  +
  \rho_1\sum_{s=1}^t\beta_1^{t-s}\xi_s .
\]
The deterministic part has magnitude at most $\mu$.  By von Bahr--Esseen,
\[
  \E\!\left[
  \left|\rho_1\sum_{s=1}^t\beta_1^{t-s}\xi_s\right|
  \right]
  \le
  C_p\sigma\rho_1^{(p-1)/p}.
\]
Therefore, for every $t>K$,
\[
  \E[u_t^+]\le\E[\abs{u_t}]
  \le
  C_p
  \left(\frac{\mu}{\sigma}+\rho_1^{(p-1)/p}\right)
  \rho_2^{(2-p)/(2p)}(\log(1/\delta))^{1/p}
  +C_\beta\delta ,
\]
where the last term uses the deterministic bound $\abs{u_t}\le C_\beta$ on
$\mathcal A_t^c$.  For the first $K$ iterates, the same deterministic bound gives
$\E[u_t^+]\le C_\beta$.  The displacement condition in the proposition is exactly
the requirement that $D_T\le R/4$, after adjusting the absolute constant $c$.
The stopped-displacement argument above completes the proof.

\end{document}